\documentclass[a4paper,twoside]{article}

\usepackage{epsfig}
\usepackage{subcaption}
\usepackage{calc}
\usepackage{amssymb}
\usepackage{amstext}
\usepackage{amsmath}
\usepackage{amsthm}
\usepackage{multicol}
\usepackage{pslatex}
\usepackage{apalike}
\usepackage[bottom]{footmisc}

\usepackage{caption}
\usepackage{subcaption}
\usepackage{adjustbox}
\usepackage{siunitx}
\usepackage{booktabs}
\usepackage{multirow}
\usepackage{csquotes}
\usepackage[dvipsnames]{xcolor}
\usepackage{nicefrac}

\usepackage[shortcuts]{extdash}

\usepackage{tikzexternal}
\tikzsetexternalprefix{figures/}

\usepackage{hyperref}
\usepackage{cleveref}
\usepackage[acronym]{glossaries}
\glsdisablehyper
\usepackage{SCITEPRESS}     % Please add other packages that you may need BEFORE the SCITEPRESS.sty package.
\begin{document}

\newacronym{nlp}{NLP}{Natural Language Processing}
\newacronym{cv}{CV}{Computer Vision}
\newacronym{mse}{MSE}{mean square error}

\title{{Number of Attention Heads} vs.\ {{Number of Transformer-Encoders} in Computer Vision}}

\author{
    \authorname{
        Tomas Hrycej\sup{1},
        Bernhard Bermeitinger\sup{1}\orcidAuthor{0000-0002-2524-1850}
        and Siegfried Handschuh\sup{1}\orcidAuthor{0000-0002-6195-9034}
    }
    \affiliation{
        \sup{1}Institute of Computer Science, University of St.Gallen (HSG), St.Gallen, Switzerland
    }
    \email{\{firstname.lastname\}@unisg.ch}
}

\keywords{%
    Computer Vision,
    Transformer,
    Attention,
    Comparison
}
\abstract{%
    Determining an appropriate number of attention heads on one hand and the number of transformer-encoders, on the other hand, is an important choice for \gls{cv} tasks using the Transformer architecture.
    Computing experiments confirmed the expectation that the total number of parameters has to satisfy the condition of overdetermination (i.e., number of constraints significantly exceeding the number of parameters).
    Then, good generalization performance can be expected.
    This sets the boundaries within which the number of heads and the number of transformers can be chosen.
    If the role of context in images to be classified can be assumed to be small, it is favorable to use multiple transformers with a low number of heads (such as one or two).
    In classifying objects whose class may heavily depend on the context within the image (i.e., the meaning of a patch being dependent on other patches), the number of heads is equally important as that of transformers.
}

\onecolumn \maketitle \normalsize \setcounter{footnote}{0} \vfill

\glsresetall
\section{\uppercase{Introduction}}\label{sec:introduction}
Architecture based on the concept of Transformers became a widespread and successful neural network framework.
Originally developed for \gls{nlp}, it has been recently also used for applications in \gls{cv}~\cite{dosovitskiy2021ImageWorth16x16}.

The key concept of a Transformer is \emph{(self\=/)~attention}.
The attention mechanism picks out segments (or words, tokens, image patches, etc.) in the input data that are building relevant context for a given segment.
This is done by means of segment weights assigned according to the similarity between the segments.
The similarity assignment can be done within multiple \emph{attention heads}.
Each of these attention heads evaluates similarity in its own way, using its own similarity matrices.
All these matrices are learned through fitting to training data.
In addition to similarity matrices, a transformer (\=/encoder) adds the results of attention heads and processes this sum through a nonlinear perceptron whose weights are also learned.
Transformer layers are usually stacked so that the output of one transformer layer is the input of the next one.
Among the most important choices for implementing a transformer-based processing system are
\begin{enumerate}
    \item
        the number of attention heads per transformer-encoder and
    \item
        the number of transformer-encoders stacked.
\end{enumerate}
The user has to select these numbers and the result substantially depends on them but it is difficult to make recommendations for these choices.
Following the general recommendation to avoid underdetermined configurations (where the number of parameters exceeds the number of constraints) and thus overfitting leading to poor generalization, there are still two above-mentioned numbers to configure: approximately the same number of network parameters can be reached by taking more attention heads in fewer transformer-encoders or vice versa.
The decision in favor of one of these alternatives may be substantial for the success of the application.
The goal of the present work is to investigate the effect of both numbers on learning performance with the help of several \gls{cv} applications.

\section{\uppercase{Parameter structure of a multi-head transformer}}\label{sec:parameter_transformer}
The parameters of a multi-head transformer (in the form of only encoders and no decoders) consist of:
\begin{enumerate}
    \item
        matrices transforming token vectors to their compressed form (\emph{value} in the transformer terminology);
    \item
        matrices transforming token vectors to the feature vectors for similarity measure (\emph{key} and \emph{query}), used for context-relevant weighting;
    \item
        matrices transforming the compressed and context-weighted form of tokens back to the original token vector length;
    \item
        parameters of a feedforward network with one hidden layer;
\end{enumerate}
All these matrices can be concatenated (e.g., column-wise) to a single parameter-vector.
Each transformer-encoder contains the same number of parameters.
The total parameter count is thus proportional to the number of transformer-encoders.
Varying the number of heads affects the parameter count resulting from the transformation matrices of the attention mechanism, the remaining ones being the parameters of the feedforward network.
The total parameter count is thus less than proportional to the number of heads.

\section{\uppercase{Measuring the degree of overdetermination}}\label{sec:measuring_overdetermination}
Fitting a parameterized structure to a data set can be viewed as an equation system.
$ M $ outputs to be fitted for $ K $ training examples constitute $ M K $ equations. $ P $ free parameters whose values are sought for the best fit correspond to $ P $ variables.
Consequently, we have a system of $ MK $ equations with $ P $ variables.
Since it is not certain that these equations can be satisfied, it is more appropriate to speak about constraints instead of equations.
In the case of linear constraints, there are well-known conditions for solvability.
Assuming mutual linear independence of constraints, this system has a unique solution if $ MK = P $.
The solution is then \emph{exactly determined}.
With $ MK < P $, the system has an infinite number of solutions --- it is \emph{underdetermined}.
In the case of $ MK > P $, the system is \emph{overdetermined} and cannot be exactly solved --- the solution is only approximate.
One such solution is based on the least-squares, i.e., minimizing the \gls{mse} of the output fit.
Usually, the real system on which the training data have been measured is assumed to correspond to a model (e.g., a linear one) with additional noise.
The noise may reflect measurement errors but also the inability of the model to describe the reality perfectly.
It is desirable that the assumed model is identified as exactly as possible while fitting the parameters to the noise in the training set is to be avoided.
The latter requirement is justified by the fact that novel patterns not included in the training set will be loaded by different noise values than those from the training set.
This undesirable fitting to the training set noise is frequently called overfitting.
For exactly determined or underdetermined configurations, the fit to the training set outputs including the noise is perfect and thus overfitting is unavoidable.
For overdetermined configurations, the degree of overfitting depends on the ratio of the number of constraints to the number of parameters.
This ratio can be denoted as
\begin{equation}\label{eq:q_ratio}
    Q = \frac{MK}{P}
\end{equation}
For a model with a parameter structure corresponding to the real system, it can be shown that the proportion of noise to which the model is fitted is equal to $ \nicefrac{1}{Q} $.
With increasing the number of training cases, this number is diminishing, with a limit of zero. Asymptotically, the \gls{mse} corresponds, in the case of white Gaussian noise, to the noise variance.
In other words, overfitting decreases with a growing number of training cases.
The dependency of \gls{mse} on the number of training samples is
\begin{equation}\label{eq:e_test}
    E
    = \sigma^2 \left( 1 - \frac{1}{Q} \right)
    = \sigma^2 \left( 1 - \frac{P}{MK} \right)
\end{equation}

The genuine goal of parameter fitting is to receive a model corresponding to the real system so that novel cases are correctly predicted.
The prediction error consists of an imprecision of the model and the noise.
For a linear regression model, the former component decreases with the size of the training set since the term ${\left( X' X \right)}^{-1} $ determines the variability of estimated model parameters (with $ X $ being the input data matrix) develops with $\nicefrac{c_1}{K}$.
The prediction is a linear combination of model parameters that amount on average to a constant $ c_2 $.
The noise component is inevitable --- its level is identical to that encountered in the training set (if both sets are  representative of the statistical population).
The resulting dependency is, with constants $ P $ and $ M $,

\begin{equation}\label{eq:e_test2}
\begin{aligned}
    E
    &= c_2 \sigma^2 {\left(X' X\right)}^{-1} + \sigma^2 \\
    &= \frac{c_1 c_2 \sigma^2}{K} + \sigma^2 \\
    &= \frac{P}{M} \frac{c_1 c_2 \sigma^2}{Q} + \sigma^2 \\
    &= \sigma^2 \left(\frac{c}{Q} + 1\right)
\end{aligned}
\end{equation}

\begin{figure*}
    \centering
    \begin{subfigure}[t]{.49\linewidth}
    \begin{adjustbox}{width=\linewidth}
        \centering
        \includegraphics[width=\linewidth]{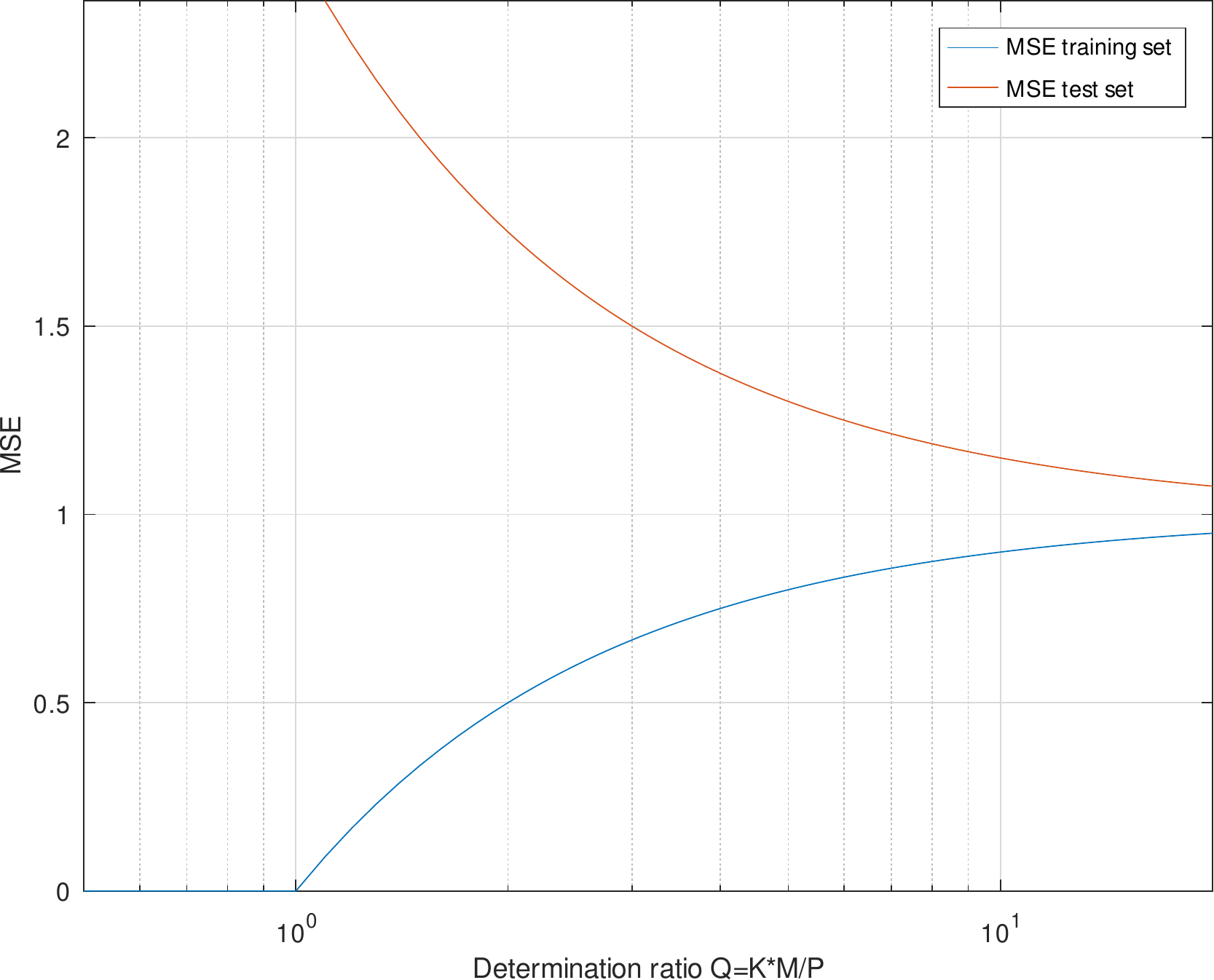}
    \end{adjustbox}
        \caption{Fixed parameter set, varying training set}\label{fig:MSE_Pconst}
    \end{subfigure}
    \hfill
    \begin{subfigure}[t]{.49\linewidth}
    \begin{adjustbox}{width=\linewidth}
        \centering
        \includegraphics[width=\linewidth]{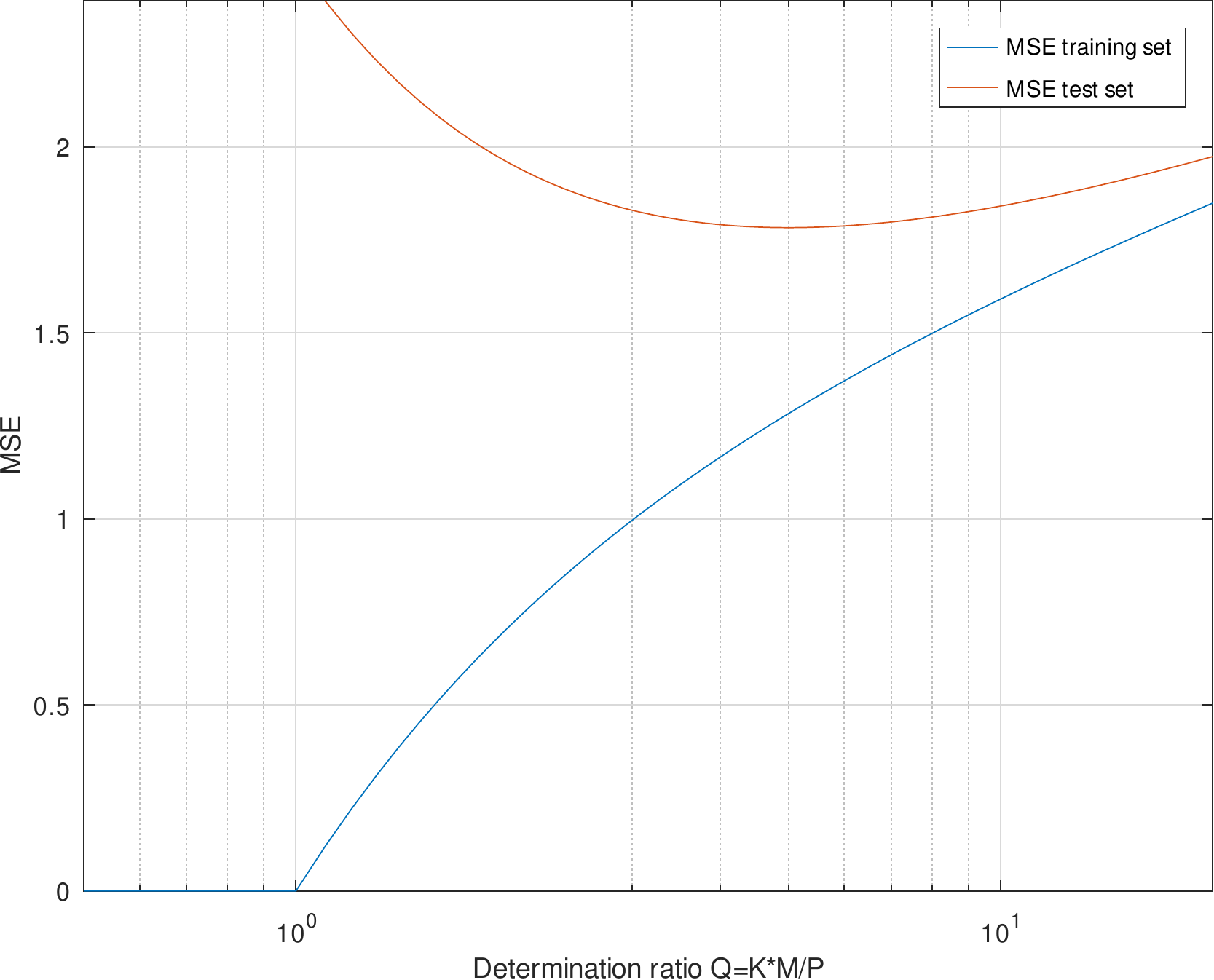}
    \end{adjustbox}
    \caption{Fixed training set, varying parameter set}\label{fig:MSE_Kconst}
    \end{subfigure}
    \vspace{0.2cm}
\caption{Training and test set \gls{mse} in dependence on determination ratio.}\label{fig:MSE_bothconst}
\end{figure*}

The shape of dependencies of training and test set \gls{mse} is exemplified in~\cref{fig:MSE_Pconst}.
The coefficient of determination on the $ x $-axis varies as the number of training samples grows.
The output dimension $ M $ and the number of model parameters $ P $ are kept constant.

In summary, the \glspl{mse} for both the training set and for the novel cases converge to the same level determined by the variance of noise if the number of training samples grows.
The condition for this is that the model structure is sufficiently expressive to capture the input/output dependence of the real system.
With nonlinear systems, these laws can be approximately justified by means of linearization.
Additionally, nonlinear systems such as layered neural networks exhibit dependencies between the parameters, the best known of which are the permutations of hidden layer units.
In the Transformer architecture, another source of redundancy are the similarity matrices of the attention mechanism.
This makes the number of genuinely free parameters $ P $ (as used above) to be below the total number of parameters.
However, the number of free parameters is difficult to assess and thus the total number can be used for a rough estimate (or as an upper bound).
Systems with $ Q > 1$ are certain to be overdetermined while those with $ Q < 1 $ are not necessarily underdetermined.
Nevertheless, the ratio $ Q $ is the best we have in practice.

\Cref{fig:MSE_Pconst} corresponds to the situation where the parameter set is kept constant while the size of the training set varies.
Frequently, the situation for choice is inverse.
There is a fixed training set and an appropriate parameter set is to be determined.
Varying (in particular, reducing) the parameter set (and maybe also the model architecture) will probably violate the condition of the model being sufficiently expressive to capture the properties of the real system.
Reducing the parameter set represents an additional source of estimation error --- the model would not be able to be perfectly fitted to training data even in the case of zero noise.
Then, the training and test set \gls{mse} will develop with an additional term growing with ratio $ Q $ (and decreasing $ P $).
The shape of this term is difficult to assess in advance without knowledge of the real system.
The typically encountered dependence is depicted in~\cref{fig:MSE_Kconst} (with arbitrary scaling of the \gls{mse}).

\section{\uppercase{Computing results}}\label{sec:computing_results}
To show the contribution of the number of heads and that of the number of transformer-encoders, a series of model fitting experiments has been performed, for several \gls{cv} classification tasks.
The data sets used have been popular collections of images, frequently used for various benchmarks.
The data sets have been chosen particularity for their match of determination ratio for the experimental networks.
Bigger data sets are deliberately left out.
For every task, a set of tasks with various pairs $ \left( h, t \right) $, the number of heads being $ h $ and the number of transformer-encoders being $ t $, have been optimized.
Some combinations with high numbers of both heads and transformer-encoders had too many parameters and have thus been underdetermined.
The consequence has been a poor test set performance.
In the following, a cross-section of the results is presented:
\begin{itemize}
    \item
        four transformer-encoders and any number of attention heads;
    \item
        four attention heads and any number of transformer-encoders.
\end{itemize}

These cross-sections contain mostly overdetermined configurations with acceptable generalization properties.
The performance has been measured by mean categorical cross-entropy on training and test sets (further referred to as \emph{loss}).
The $ x $-axis of~\cref{fig:MSE_bothconst,fig:mnist_32_64_4,fig:cifar100_64_128_4,fig:birds_128_32_4,fig:places_128_32_4,fig:imagenet_128_64_4} is the determination ratio $ Q $ of~(\cref{eq:q_ratio}), in logarithmic scale (so that the value $10^0$ corresponds to $ Q = 1 $).
This presentation makes the dependence of the generalization performance (as seen in the convergence of the training and the test set cross-entropy) on the determination ratio clear.
This ratio grows with the decreasing number of parameters, that is, with the decreasing number of heads if the number of transformer-encoders is kept to four and the decreasing number of transformer-encoders if the number of heads is kept to four.
The rightmost configuration is that with a single head or a single transformer-encoder, respectively, followed to the left with two heads or two transformer-encoders, etc..

The optimization was done exclusively with single precision (\textit{float32}) over a fixed number of \num{100} epochs by \emph{AdamW}~\cite{loshchilov2019DecoupledWeightDecay} with a learning rate of \num{1e-3} and a weight decay of \num{1e-4}.
For consistency, the batch size was set to \num{256} for all experiments.

As a simple regularization during training, standard image augmentation techniques were applied: random translation by a factor of $ (0.1, 0.1) $, random rotation by a factor of \num{0.2}, and random cropping to \SI{80}{\percent}.

The patches are flattened and their absolute position is added in embedded form to each patch before entering the first encoder.

All experiments were individually conducted on one \emph{Tesla V100-SXM3-32GB} GPU for a total number of 60 GPU days.

\subsection{Dataset MNIST}\label{sec:dataset_mnist}
\begin{figure}
    \centering
    \tikzsetnextfilename{mnist-32-64.pdf}
    \begin{tikzpicture}
        \centering
        \begin{semilogxaxis}[
            default,
            width=\columnwidth,
            legend pos=north west,
            xlabel={Determination ratio $ Q = \nicefrac{KM}{P} $},
            ylabel={Loss},
        ]
            \addplot[
                solid, color=RoyalBlue, mark=x,
            ] table [
                x=determination, y=loss
            ] {data/mnist-32-64_t4.data};
            \addlegendentry{$ \text{TF} = 4 $ train}

            \addplot[
                dashed, color=RoyalBlue, mark=o,
            ] table [
                x=determination, y=val_loss
            ] {data/mnist-32-64_t4.data};
            \addlegendentry{$ \text{TF} = 4 $ test}

            \addplot[
                solid, color=BrickRed, mark=x,
            ] table [
                x=determination, y=loss
            ] {data/mnist-32-64_h4.data};
            \addlegendentry{$ \text{HD} = 4 $ train}

            \addplot[
                dashed, color=BrickRed, mark=o,
            ] table [
                x=determination, y=val_loss
            ] {data/mnist-32-64_h4.data};
            \addlegendentry{$ \text{HD} = 4 $ test}
        \end{semilogxaxis}
    \end{tikzpicture}
    \vspace{0.2cm}
    \caption{Training and test set losses of model variants for dataset \emph{MNIST}.}\label{fig:mnist_32_64_4}
\end{figure}
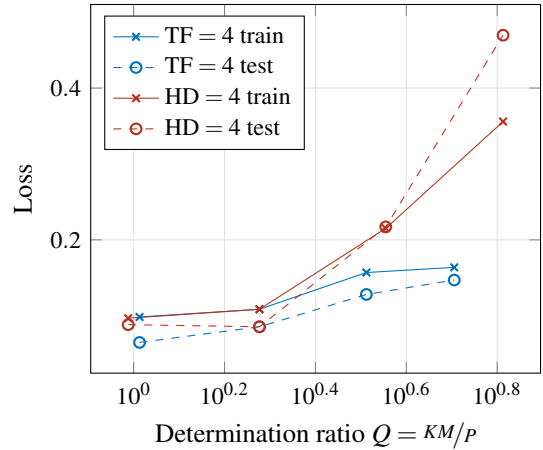

The MNIST~\cite{lecun1998GradientbasedLearningApplied} dataset consists of pixel images of digits.
All pairs $ \left( h, t \right) $ with number of heads $ h \in \{1, 2, 4, 8\} $  and number of transformer-encoders $ t \in \{1, 2, 4, 8\} $ have been optimized.
The results in the form of loss depending on the determination ratio $ Q $ are given in~\cref{fig:mnist_32_64_4}.

The gray-scale images were resized to $ 32 \times 32 $ and the patch size was set to \num{2}.
All internal dimensions (keys, queries, values, feedforward, and model size) are set to \num{64}.

The cross-entropies for the training and the test sets are fairly consistent, due to the determination ratio $ Q > 1 $.
The results are substantially more sensitive to the lack of transformer-encoders: the rightmost configurations with four heads but one or two transformer-encoders have a poor performance.
By contrast, using only one or two heads leads only to a moderate performance loss.
In other words, it is more productive to stack more transformer-encoders than to use many heads.
This is not surprising for simple images such as those of digits.
The context-dependency of image patches can be expected to be rather low and to require only a simple attention mechanism with a moderate number of heads.

\subsection{Dataset CIFAR-100}\label{sec:dataset_cifar100}
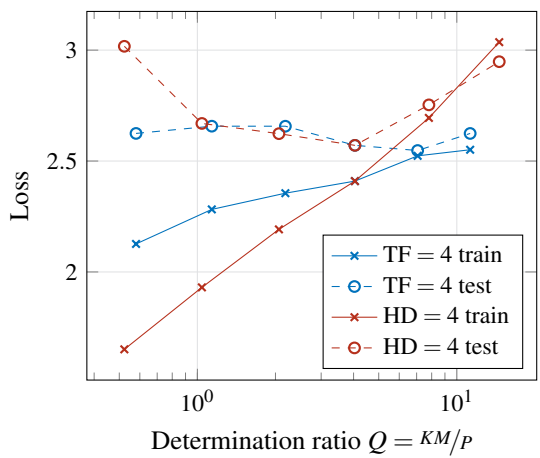
\begin{figure}
    \centering
    \tikzsetnextfilename{cifar100-64-128.pdf}
    \begin{tikzpicture}
        \centering
        \begin{semilogxaxis}[
            default,
            width=\columnwidth,
            legend pos=south east,
            xlabel={Determination ratio $Q = \nicefrac{KM}{P}$},
            ylabel={Loss},
        ]
            \addplot[
                solid, color=RoyalBlue, mark=x,
            ] table [
                x=determination, y=loss
            ] {data/cifar100-64-128_t4.data};
            \addlegendentry{$ \text{TF} = 4$ train}

            \addplot[
                dashed, color=RoyalBlue, mark=o,
            ] table [
                x=determination, y=val_loss
            ] {data/cifar100-64-128_t4.data};
            \addlegendentry{$ \text{TF} = 4$ test}

            \addplot[
                solid, color=BrickRed, mark=x,
            ] table [
                x=determination, y=loss
            ] {data/cifar100-64-128_h4.data};
            \addlegendentry{$ \text{HD} = 4$ train}

            \addplot[
                dashed, color=BrickRed, mark=o,
            ] table [
                x=determination, y=val_loss
            ] {data/cifar100-64-128_h4.data};
            \addlegendentry{$ \text{HD} = 4$ test}
        \end{semilogxaxis}
    \end{tikzpicture}
    \vspace{0.2cm}
    \caption{Training and test set losses of model variants for dataset \emph{CIFAR-100}.}\label{fig:cifar100_64_128_4}
\end{figure}

The dataset \emph{CIFAR-100}~\cite{krizhevsky2009LearningMultipleLayers} is a collection of images of various object categories such as animals, household objects, buildings, people, and others.
The objects are labeled into \num{100} classes.
The training set consists of \num{50 000}, the test set of \num{10 000} samples.
With $ M = 100 $ and $ K = 50\,000$, the determination coefficient $ Q $ is equal to unity (100 on the plot x-axis) for 5 million free parameters ($M \times K$). The results are given in~\cref{fig:cifar100_64_128_4}.
All pairs $ (h,t) $ with number of heads $ h \in \{1, 2, 4, 8, 16, 32\} $ and number of transformer-encoders $ t \in \{1, 2, 4, 8, 16, 32\} $ have been optimized.

The colored images were up-scaled to $ 64 \times 64 $ and the patch size was set to \num{8}.
All internal dimensions (keys, queries, values, feedforward, and model size) are set to \num{128}.

The cross-entropies for the training and the test sets converge to each other for about $ Q > 4 $, with a considerable generalization gap for $ Q < 1 $.
This can be expected taking theoretical considerations mentioned in~\cref{sec:measuring_overdetermination} into account.
The results are more sensitive to the lack of transformer-encoders than to that of heads.
How far a high number of transformer-encoders would be helpful, cannot be assessed because of getting then into the region of $ Q < 1 $.
With this training set size, a reduction of some transformer parameters such as key, query, and value width would be necessary.

\subsection{Dataset CUB-200-2011}\label{sec:dataset_birds}
\begin{figure}
    \centering
    \tikzsetnextfilename{birds-128-32.pdf}
    \begin{tikzpicture}
        \centering
        \begin{semilogxaxis}[
            default,
            width=\columnwidth,
            legend pos=north west,
            xlabel={Determination ratio $ Q = \nicefrac{KM}{P} $},
            ylabel={Loss}
        ]
            \addplot[
                solid, color=RoyalBlue, mark=x,
            ] table [
                x=determination, y=loss
            ] {data/birds-128-32_t4.data};
            \addlegendentry{$ \text{TF} = 4 $ train}

            \addplot[
                dashed, color=RoyalBlue, mark=o,
            ] table [
                x=determination, y=val_loss
            ] {data/birds-128-32_t4.data};
            \addlegendentry{$ \text{TF} = 4$ test}

            \addplot[
                solid, color=BrickRed, mark=x,
            ] table [
                x=determination, y=loss
            ] {data/birds-128-32_h4.data};
            \addlegendentry{$ \text{HD} = 4$ train}

            \addplot[
                dashed, color=BrickRed, mark=o,
            ] table [
                x=determination, y=val_loss
            ] {data/birds-128-32_h4.data};
            \addlegendentry{$ \text{HD} = 4$ test}
        \end{semilogxaxis}
    \end{tikzpicture}
    \caption{Training and test set losses of model variants for dataset \emph{birds}.}\label{fig:birds_128_32_4}
\end{figure}
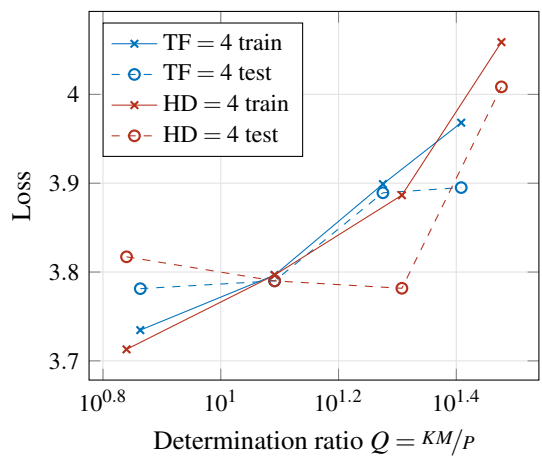

The training set of the dataset \emph{CUB-200-2011}~\cite{wah2011CaltechUCSDBirds2002011Dataset} (\emph{birds}) used for the image classification task consists of \num{5 994} images of birds of \num{200} species.
All pairs $(h,t)$ with number of heads $ h \in \{1, 2, 4, 8\} $ and number of transformer-encoders $ t \in \{1, 2, 4, 8\} $ have been optimized (\cref{fig:birds_128_32_4}).

The colored images were resized to $ 128 \times 128 $ and the patch size was set to \num{8}.
All internal dimensions (keys, queries, values, feedforward, and model size) are set to \num{32}.

The cross-entropies for the training and the test sets are mostly consistent due to the high determination ratio $ Q $.
There are relatively small differences between small numbers of heads and transformer-encoders.
Both categories seem to be comparable.
This suggests, in contrast to the datasets treated above, a relatively large contribution of context to the classification performance --- multiple heads are as powerful as multiple transformer-encoders.
This is not unexpected in the given domain: the habitat of the bird in the image background may constitute a key contribution to classifying the species.

\subsection{Dataset places365}\label{sec:dataset_places}
\begin{figure}
    \centering
    \tikzsetnextfilename{places-128-32.pdf}
    \begin{tikzpicture}
        \centering
        \begin{semilogxaxis}[
            default,
            width=\columnwidth,
            legend pos=south east,
            xlabel={Determination ratio $Q = \nicefrac{KM}{P}$},
            ylabel={Loss},
            ymin=3.5,
        ]
            \addplot[
                solid, color=RoyalBlue, mark=x,
            ] table [
                x=determination, y=loss
            ] {data/places-128-32_t4.data};
            \addlegendentry{$ \text{TF} = 4$ train}

            \addplot[
                dashed, color=RoyalBlue, mark=o,
            ] table [
                x=determination, y=val_loss
            ] {data/places-128-32_t4.data};
            \addlegendentry{$ \text{TF} = 4$ test}

            \addplot[
                solid, color=BrickRed, mark=x,
            ] table [
                x=determination, y=loss
            ] {data/places-128-32_h4.data};
            \addlegendentry{$ \text{HD} = 4$ train}

            \addplot[
                dashed, color=BrickRed, mark=o,
            ] table [
                x=determination, y=val_loss
            ] {data/places-128-32_h4.data};
            \addlegendentry{$ \text{HD} = 4$ test}
        \end{semilogxaxis}
    \end{tikzpicture}
    \caption{Training and test set losses of model variants for dataset \emph{places}.}\label{fig:places_128_32_4}
\end{figure}
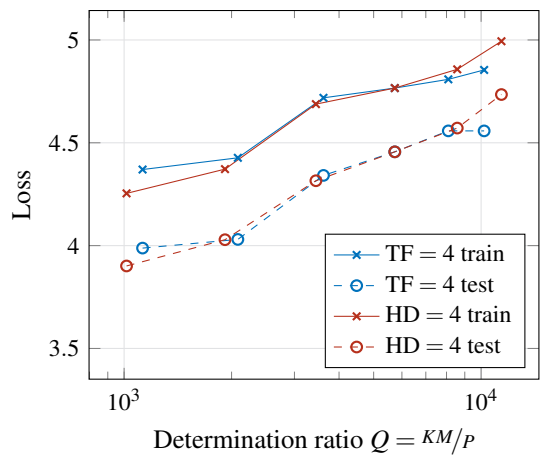

The training set of dataset \emph{places365}~\cite{zhou2018Places10Million} consists of \num{1 803 460} images of various places in \num{365}~classes (\cref{fig:places_128_32_4}).
Pairs $ \left( h, t \right) $ with number of heads $ h \in \{1, 2, 4, 8, 16, 32\} $ and number of transformer-encoders $ t \in \{1, 2, 4, 8, 16, 32\} $ have been optimized.

The colored images were resized to $ 128 \times 128 $ and the patch size was set to \num{16}.
All internal dimensions (keys, queries, values, feedforward, and model size) are set to \num{32}.

The cross-entropies for the training and the test sets are parallel.
Surprisingly, test set losses are lower than those for the training set.
This can be caused by an inappropriate test set containing only easy-to-classify samples.
The reason for this training to test consistency is the very high determination ratio $ Q $
(over \num{1 000}). This would allow even larger numbers of transformer-encoders and heads without worry about generalization, with a corresponding high computing expense.

There are hardly any differences between variants with varying heads and those varying transformer-encoders.
With a given total number of parameters (and thus a similar ratio $Q$), both categories seem to be equally important.
It can be conjectured that there is a relatively strong contribution of context to the classification performance can be assumed.

\subsection{Dataset imagenet}\label{sec:dataset_imagnet}
The training set of the popular \emph{imagenet}~\cite{krizhevsky2012ImageNetClassificationDeep} dataset contains \num{1 281 167} images of \num{1 000} different classes of current everyday objects (like airplanes, cars, different types of animals, etc.)

For this dataset, the pairs $ \left( h, t \right) $ with number of heads $ h \in \{1, 2, 4, 8\} $ and number of transformer-encoders $ t \in \{1, 2, 4, 8\} $ have been optimized.

Analog to the \emph{places} experiment, the colored images were resized to $ 128 \times 128 $ and the patch size was set to \num{16}.
All internal dimensions (keys, queries, values, feedforward, and model size) are set to \num{64}.

\begin{figure}
    \centering
    \tikzsetnextfilename{imagenet-128-64.pdf}
    \begin{tikzpicture}
        \centering
        \begin{semilogxaxis}[
            default,
            width=\columnwidth,
            legend pos=north west,
            xlabel={Determination ratio $Q = \nicefrac{KM}{P}$},
            ylabel={Loss}
        ]
            \addplot[
                solid, color=RoyalBlue, mark=x,
            ] table [
                x=determination, y=loss
            ] {data/imagenet-128-64_t4.data};
            \addlegendentry{$ \text{TF} = 4$ train}

            \addplot[
                dashed, color=RoyalBlue, mark=o,
            ] table [
                x=determination, y=val_loss
            ] {data/imagenet-128-64_t4.data};
            \addlegendentry{$ \text{TF} = 4$ test}

            \addplot[
                solid, color=BrickRed, mark=x,
            ] table [
                x=determination, y=loss
            ] {data/imagenet-128-64_h4.data};
            \addlegendentry{$ \text{HD} = 4$ train}

            \addplot[
                dashed, color=BrickRed, mark=o,
            ] table [
                x=determination, y=val_loss
            ] {data/imagenet-128-64_h4.data};
            \addlegendentry{$ \text{HD} = 4$ test}
        \end{semilogxaxis}
    \end{tikzpicture}
    \caption{Training and test set losses of model variants for dataset \emph{imagenet}.}\label{fig:imagenet_128_64_4}
\end{figure}
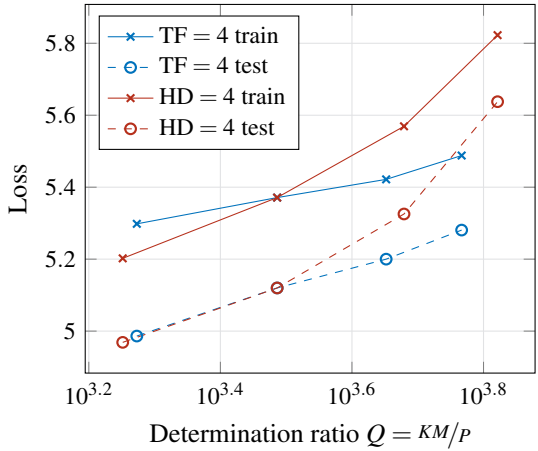

For this experiment, it can be seen in the determination ratios in~\cref{fig:imagenet_128_64_4} that it behaves similarly to the dataset \emph{places} (in~\cref{fig:places_128_32_4}).
Again, the test loss is consistently lower than the training loss.
The lowest cross-entropies are comparable which means, analog to \emph{places}, that increasing the number of attention heads and the number of transformer-encoder layers is beneficial to the performance.
Compared to the other experiments, the determination ratio is very high (\numrange{e3}{e4}) which means that the number of parameters in the classification network is too small and even larger stacks of transformer-encoders with more attention heads could decrease the loss even further.

Looking at the varying number of attention heads, it can be seen that their number has a low impact on the performance.

\section{Conclusions}\label{sec:conclusions}
Determining the appropriate number of self-attention heads on one hand and, on the other hand, the number of transformer-encoder layers is an important choice for \gls{cv} tasks using the Transformer architecture.

A key decision concerns the total number of parameters to ensure good generalization performance of the fitted model.
The determination ratio $ Q $, as defined in~\cref{sec:measuring_overdetermination}, is a reliable measure: values significantly exceeding unity (e.g., $ Q > 4 $) lead to test set loss similar to that of the training set.
This sets the boundaries within which the number of heads and the number of transformer-encoders can be chosen.

Different \gls{cv} applications exhibit different sensitivity to varying and combining both numbers.
\begin{itemize}
\item
    If the role of context in images to be classified can be assumed to be small, it is favorable to \enquote{invest} the parameters into multiple transformer-encoders.
    With too few transformer-encoders, the performance will rapidly deteriorate.
    Simultaneously, a low number of attention heads (such as one or two) is sufficient.
\item
    In classifying objects whose class may heavily depend on the context within the image (i.e., the meaning of a patch being dependent on other patches), the number of attention heads is equally important as that of transformer-encoders.
\end{itemize}

This seems to be consistent with other experiments like~\cite{li2022StructuralAttentionEnhanced} where the optimal number of attention heads depends on the dataset.

\paragraph{Future work}
Although this study provides a systematic comparison between the number of attention heads and number of consecutive transformer-encoders, the sheer number of different hyperparameters is still underrepresented.
The hyperparameters in this study were chosen for the task at hand, e.g.\ the patch size was chosen accordingly to the input image size.
However, the patch size is on its own a crucial hyperparameter which might lead to different results if chosen differently.
Any of the listed hyperparameters in the experiments (\cref{sec:computing_results}) need the same systematic analysis as the current study.
This is left out for future work.

\bibliographystyle{apalike}
{\small \bibliography{bibliography}}

\end{document}